# Natural Scene Text Editing Based on AI

Yujie Zhang*

Beijing University of Chemical Technology, ChaoYang District, Beijing, People's Republic of China 100029
* Corresponding author: zhangyujie719@163.com

**ABSTRACT**

In natural photography, text editing is used to replace or update a word in an image while keeping the image's authenticity. The ability to edit text directly on photographs has a number of benefits, including the ability to correct mistakes and restore text. This is tough since the background and text styles of the altered image must match those of the original. This research shows how to change image text at the letter and digits level. I devised a two-part letters-digits network (LDN) to encode and decode digital images, as well as learn and transfer the font style of the source characters to the target characters. This method allows you to update the uppercase letters, lowercase letters and digits in the picture.

## 1. INTRODUCTION

A variety of scenario imagers and design software tools might identify letters and numbers. Changing the text in an image is tricky for a variety of reasons. It is challenging to create unseen characters with appropriate visual coherence given the tiny amount of viewable characters.[1] The goal of this post is to keep the background while editing the uppercase, lowercase, and digits (62-letters) in the image. When updating text, I need to preserve the original background. It was exhibited a context-based pixel prediction-driven unsupervised visual feature learning technique. Context encoders can be employed on their own or as part of a non-parametric semantic inpainting method's setup. A context encoder learns a representation of visual structures that includes both the look and the meaning of the structures.[2] In natural language processing, self-supervised pretraining has effectively addressed data thirst. Today's models may rapidly outgrow one million photographs and require hundreds of millions of annotated images, which are typically inaccessible to the general public. Deep learning has resulted in an explosion of architectures (such as NLP) with ever-increasing capabilities and capacity. The training of generalizable NLP models with over a hundred billion parameters is now possible. In GP, the solutions are based on autoregressive language modeling, while in BERT, the solutions are based on masked autoencoding. A fundamental method that works well in computer vision is to hide a large number of random patches to facilitate learning favorable features. This method drastically lowers redundancy while enforcing a high level of self-supervision. MAE subtracts random patches from the input picture and reconstructs them in pixel space. It has an asymmetric encoder-decoder architecture and light hardware. Experiments were carried out on a data set of 10,000 typefaces in a variety of styles. It was suggested that an end-to-end stacked conditional GAN model be utilized, which takes into account material across channels. From a small number of instances, a collection of multi-content photographs with a consistent look was created.[3]

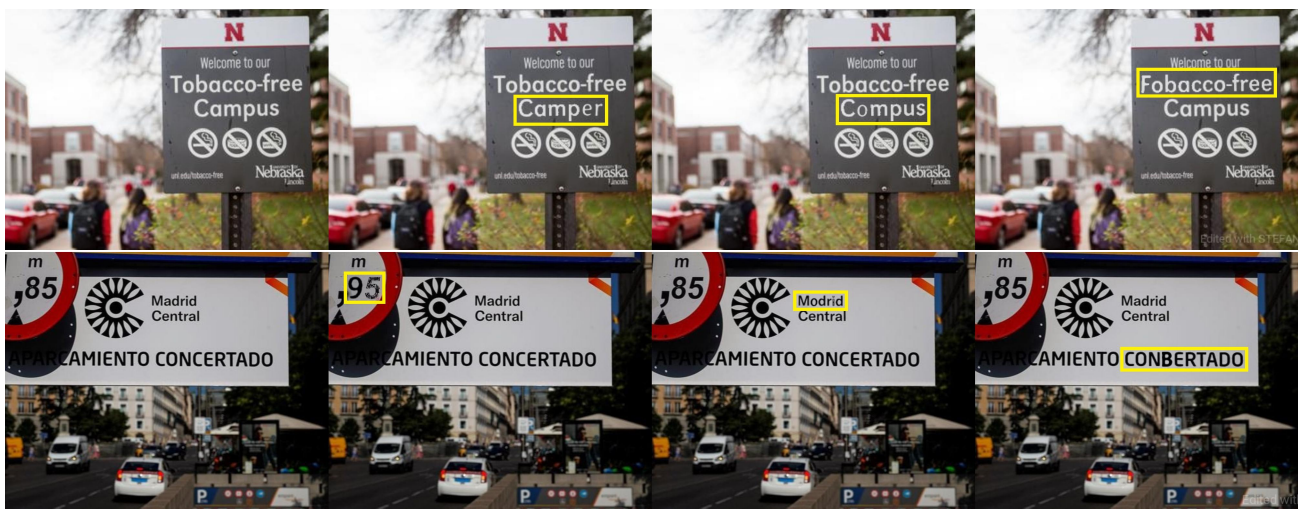

Figure 1. The task is depicted qualitatively. Automatic editing utilizing our letter-number encoder training is displayed on the right, given an image on the left containing some letter or number.

## 2. PREVIOUS WORK

### 2.1 Font Style Transfer

Many design and scene images include text, which provides visitors with important context information. Due to the limited number of visible characters, creating invisible characters with adequate visual consistency is difficult. Natural characteristics like brightness, contrast, shadow, perspective distortion, and others make it more difficult to replace a character straight from a picture. For a long time, the problem of creating typeface glyphs from a few examples has been studied. Previous approaches relied heavily on explicit form modeling to create the transition between existing and novel glyphs. As a result of the rise of deep learning, convolutional neural networks have recently been utilized to produce new glyphs. Different from inferring the general glyph design, transferring creative styles of color and texture to new glyphs is a difficult task.[4] Neural networks are used to train computers to read text in a variety of typefaces. A font's style is learned by studying a subset of only four letters.[5] Computer graphics with a dissimilarity to reality of 22.4 percent less than state-of-the-art are developed. The task of developing a 62-letter font is solved using an innovative neural network design for dealing with single-image analogies.[6] For the ornamented font, an end-to-end stacked conditional GAN model that considers content along channels and style throughout network layers is provided.[7] A large-scale benchmark in a number of styles can be constructed and rendered in a range of typefaces to represent the human hand's gestures and writing style, among other things.[8] A suggested method for creating Chinese handwritten typefaces that successfully reuses sample characters supplied by users. The prototype only requires users to enter around 20% of the characters, as is customary.[9] The framework for learning to solve these problems is provided by bilinear models, which include spoken and non-verbal representations of sound and sight.[10] The forged-and-recaptured samples created by the proposed text-editing attack and recapturing procedure have fooled several existing document authentication systems.[11] Post-processing can be used to transform the glyphs into vector typefaces. Glyphs are intended to be represented as form primitives contained by quadratic curves, resulting in a single implicit glyph shape representation.[12] Both metric learning and image forensic approaches are used in the suggested algorithm.[13] In 3D computer graphics, textures that are true to semantic classes may be obtained. A novel Spatial Feature Transform (SFT) layer creates affine transformation parameters for spatially-wise feature modulations.[14]

### 2.2 Scene Text Detector

EP seeks to estimate the likelihood of constructing a string from the probability distribution's output sequence. Characters that are lacking, superfluous, or unknown may be the focus of the training process.[15] Using just line-level text-images, it enables for completely autonomous and unsupervised learning. On both synthetically made text pictures and scanned photos of real-printed books, text recognition accuracy is excellent.[16] A new technique has been developed to produce 3D textures that can match and adapt to local conditions. It employs a hybrid of generative Markov random field (MRF) models with deep convolutional neural networks.[17] As part of an approach for identifying scene text, the corner points of text bounding boxes are found, and text areas are split in relative positions. The approach handles lengthy orientated text without the need for additional post-processing.[18] A pipeline has been devised that predicts words or text lines of various orientations and quadrilateral configurations.[19] A novel architecture is being created that takes advantage of convolutional neural networks' (CNN) powerful capabilities to handle some of the most difficult tasks in image-based text recognition, such as text-to-speech translation and e-text coordination.[20] Project Naptha analyzes photographs and assists you in finding essential terms in web material, as well as highlighting, copying, and pasting text from an image into a text editor.[21] To solve the challenge of lexicon-free text extraction from complex texts, an end-to-end trainable multi-task network is deployed. With no post-processing, chopping, or word grouping, this network handles both text localization and text recognition difficulties concurrently. In uncommon OCR circumstances, the model outperformed benchmark datasets and comparable techniques.[22]

## 3. METHODOLOGY

I came up with a two-part letters-digits network (LDN). (1) Remove the source characters from the original picture and restore the background through the encoder-decoder pipline (2) Learn the font style of the source characters and transfer the style to the target characters. Place the processed target character in the restored background.

### 2.3 Pipeline for Encoders and Decoders

A fully-connected layer connects an encoder and a decoder, allowing each unit in the decoder to reason about the complete picture content. The general design is a straightforward encoder-decoder pipeline. The encoder creates a latent feature representation of an image with missing areas from an input picture. The decoder uses this feature representation to generate the missing picture material, which may then be analyzed. Context Encoder is a program that translates a situation's context. The encoder gets the context image and extracts features, which are subsequently sent to the decoder through a fully-connected channel layer. The decoder then reconstructs the image's missing components.

### 2.4 Font Style Migration Network

An end-to-end network can predict the full set of stylistically related photographs based on a subset of styled photos from a certain category. The first network predicts glyph masks, while the second network fine-tunes the colour and ornamentation of the final glyphs. By training on the 10K font dataset and generalizing to glyph prediction for any arbitrary font given a few grayscale letters, the glyph network learns the overall structure of the "font manifold." The ornamentation network fine-tunes the color and ornamentation of these coarse glyph shapes for each randomly generated typeface, resulting in clean, well-styled letters. The issue is divided into digestible pieces, each with its own loss on the intermediate output and a progressive training strategy that pre-trains GlyphNet before fine-tuning the entire thing for improved regularization.

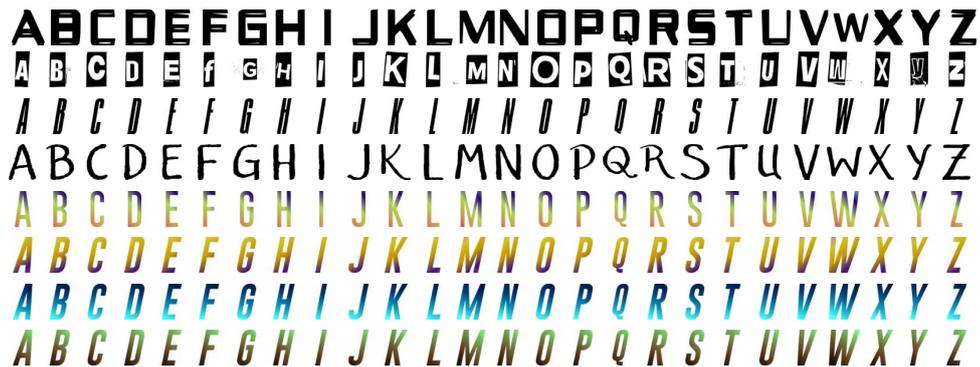

Figure 2. A subset of the 20K grayscale and color font dataset was chosen at random. To produce a foundation collection of upgraded fonts, a 20K color font data set is built by utilizing random color gradients and highlighting on grayscale glyphs. This data collection's size may be arbitrarily enlarged by producing additional random colors. These gradient typefaces do not have the same distribution as natural ornamentations, but they may be utilized for network pre-training applications.[4]

# 4. CONCLUSION

Vision in computers Self-supervised learning may be heading in the same direction as natural language processing. Rather than entities, the MAE reconstructs distinct types of signals. GANs can help with the problem of multi-content style transfer. Because two sub-networks are trained: one for shape and one for texture, the findings outperform prior glyph-focused texture transfer techniques. It explains how to use GANs to solve a variety of content-style transfer problems.

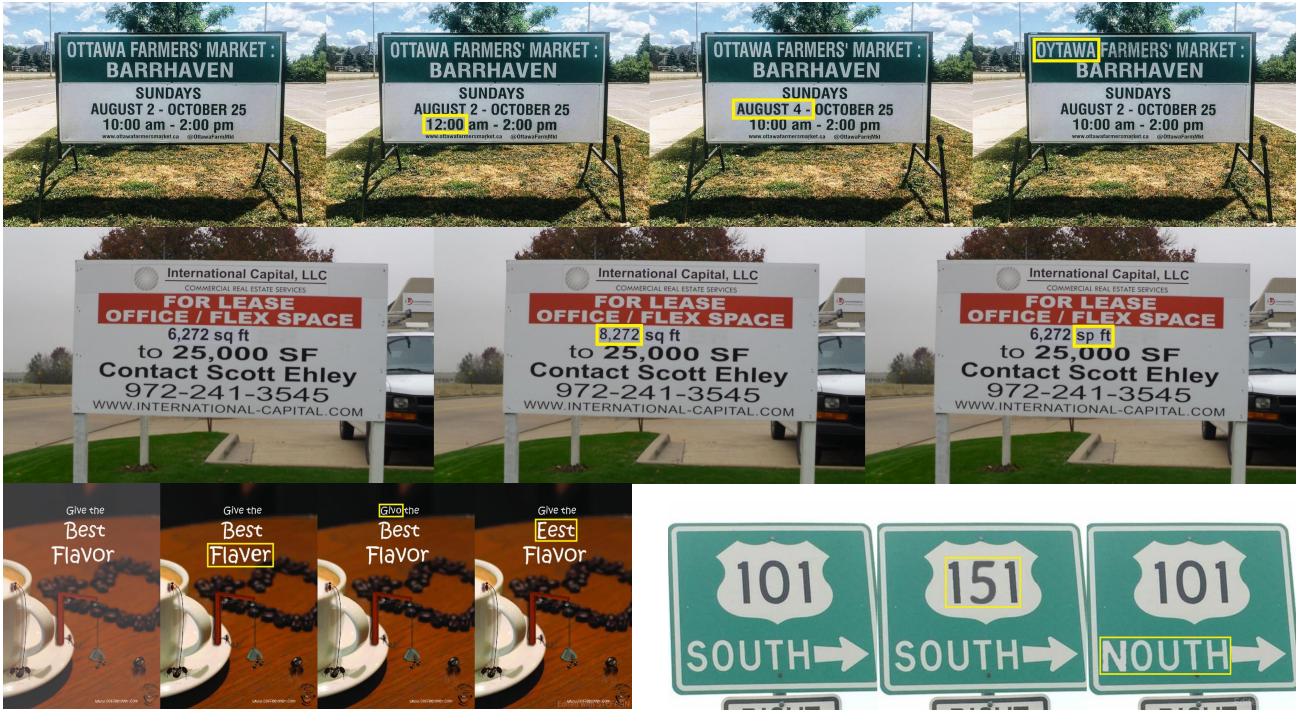

Figure 3. Examples of text editing with a letters-digits network (LDN): The photographs on the left are the original images from the dataset, while the images on the right have been edited. As seen, LDN can change multiple characters in a word (uppercase letters, lowercase letters, and numerals) as well as an entire word in a text region.

The primary purpose of a letter-digit network (LDN) is to do image editing for a variety of applications such as error correction, text restoration, image reusability, and others. LDN alters scene text for a single or several characters (uppercase, lowercase, and numerals) to guarantee visual consistency and inter-character spacing. As indicated by the outputs, LDN can successfully detect the text in the image, keep the background texture, and modify the uppercase letters, lowercase letters, and numbers. The method yielded impressive subjective visual realism findings on random real-world pictures. Future goals include developing a new approach to evaluating text editing quality, transferring information across different language combinations, continuing to train LDN to improve its intensive reading of text style conversion, and focusing on text editing in more difficult contexts involving multiple languages. Also, to avoid misusing the natural text editing network, devise methods for detecting processed photos, such as smearing the modified text region or quantitatively quantifying the degree of image change.